\documentclass[final,5p,times,twocolumn]{elsarticle}

\usepackage[english]{babel}
\usepackage{amssymb, amsmath}
\usepackage{amsmath,amsfonts,amssymb,amsthm,epsfig,url,epstopdf,array}
\usepackage{graphicx}
\usepackage[justification=centering]{caption}
\usepackage{multirow}
\usepackage{enumerate}
\usepackage{color}
\usepackage[pdftex,dvipsnames,table]{xcolor}
\usepackage{float}
\usepackage{subcaption}
\usepackage[flushleft]{threeparttable}
\usepackage{booktabs,caption}
\usepackage[belowskip=-7pt,aboveskip=5pt]{caption}
\usepackage{xargs}
\usepackage{pgf}
\usepackage{tikz}
\usepackage{tabularx}
\usepackage{relsize}
\usetikzlibrary{arrows,automata,positioning}
\usepackage[flushleft]{threeparttable} 
\usepackage{algorithm}
\usepackage{algpseudocode}
\usepackage{color, colortbl}
\definecolor{Gray}{gray}{0.9}
\definecolor{lightcyan}{rgb}{0.92, 1, 1}
\definecolor{aliceblue}{rgb}{0.94, 0.97, 1.0}
\usepackage{threeparttable}

\journal{Pattern Recognition Letters}

\begin{document}

\begin{frontmatter}

\title{A fuzzy-rough uncertainty measure to discover bias encoded explicitly or implicitly\\ in features of structured pattern classification datasets}

\author[TILBURG]{Gonzalo N\'apoles\corref{mycorrespondingauthor}%
}
\ead{g.r.napoles@uvt.nl}
\address[TILBURG]{Department of Cognitive Science \& Artificial Intelligence, Tilburg University, The Netherlands.}

\author[HASSELT]{Lisa Koutsoviti Koumeri}
\ead{lisa.koutsoviti@uhasselt.be}
\address[HASSELT]{Business Informatics Research Group, Hasselt University, Belgium.}

\cortext[mycorrespondingauthor]{Corresponding author}

\begin{abstract}

The need to measure bias encoded in tabular data that are used to solve pattern recognition problems is widely recognized by academia, legislators and enterprises alike. In previous work, we proposed a bias quantification measure, called fuzzy-rough uncertainty, which relies on the fuzzy-rough set theory. The intuition dictates that protected features should not change the fuzzy-rough boundary regions of a decision class significantly. The extent to which this happens is a proxy for bias expressed as uncertainty in a decision-making context. Our measure's main advantage is that it does not depend on any machine learning prediction model but a distance function. In this paper, we extend our study by exploring the existence of bias encoded implicitly in non-protected features as defined by the correlation between protected and unprotected attributes. This analysis leads to four scenarios that domain experts should evaluate before deciding how to tackle bias. In addition, we conduct a sensitivity analysis to determine the fuzzy operators and distance function that best capture change in the boundary regions.
\end{abstract}


\end{frontmatter}


\section{Introduction}
\label{sec:introduction}

Data-driven decision support systems have been accused of being a fertile ground to produce biased results, thus leading to discriminatory decisions~\cite{balayn2021managing}. As historical data often encode biases~\cite{Fuchs} explicitly or implicitly~\cite{elsbach2019new}, pattern recognition algorithms inevitably relate their predictions with protected characteristics such as race or gender. The Equality Act 2010 of government of the United Kingdom defines protected attributes as personal characteristics (such as gender or race) that should not put a person at a substantial disadvantage compared to people with different personal characteristics. In the literature, more than 20 definitions of fairness~\cite{Ntoutsi} and respective bias metrics have been proposed. However, existing metrics express different and often contradictory notions of fairness~\cite{Verma,Corbett,Friedler} depending on local legal and cultural conventions~\cite{hajian2012methodology} or on the type of decision-support system~\citep{balayn2021managing}. Deciding which metric is most appropriate for the task at hand is difficult~\cite{Kusner} as several parameters need to be considered such as causal influences among features, mis-representation of groups and different modalities of data~\cite{Ntoutsi}. Therefore, the need for introducing general-purpose, direct and indirect bias measures is evident~\cite{hajian2012methodology}.

The literature on this subject relies on two dominating notions of fairness: group-based and individual-based fairness measures~\cite{binns2020apparent}. Group-based measures have been criticized for leading to inverse discrimination~\cite{Chouldechova} and being oblivious to features other than the sensitive feature~\cite{Corbett, Choi}. Moreover, they often require discretization of numeric sensitive features such as age, which can alter bias measures' outputs~\cite{Friedler}. Individual-based fairness measures require strong assumptions such as the availability of an agreed-upon similarity metric, or knowledge of the underlying data generating process~\cite{Kehrenberg}. These measures act as bias proxies as they do not measure bias directly. For example, they can rely on the consistency in classification or the redundancy in data. Finally, both groups of measures are often applied on predictions generated by black-box machine learning models for fairness assessment~\cite{aif3602018}. However, most successful prediction models are not intuitively explainable~\cite{holstein2019improving} and tend to be sensitive to variations in the input arising from variations in training-test splits~\cite{Friedler}.

Another sensitive issue refers to implicit bias or indirect discrimination, which occurs when decisions are made based on nonsensitive features that strongly correlate with biased sensitive ones~\cite{hajian2012methodology}. This means that even if protected features are excluded from the decision making process, a classification algorithm might still produce biased results. Existing implicit bias measures are found in~\cite{pedreshi2008discrimination, hajian2012methodology} where background knowledge is used to manually set classification rules combined with discriminatory thresholds. The possible pitfall in such an approach is that human experts might misjudge the impact of feature categories on the decision outcomes~\citep{grgic2018beyond}.

Recently, we proposed a measure called fuzzy-rough uncertainty (FRU) to quantify explicit bias of protected features in pattern classification problems~\cite{koutsoviti2021bias}. Our measure quantifies the changes in the fuzzy-rough boundary regions after removing a protected feature as a proxy for measuring fairness. To that end, we use the advantages of rough sets~\cite{Pawlak} for analyzing inconsistency in decision systems. Measuring the distance or the change between the regions of fuzzy-rough sets has been examined in the literature~\cite{fuzz_dist, Bello}, but not in the context of bias quantification, as far as we know. To cope with the issue of defining similarity thresholds when handling problems involving continuous features, we use fuzzy-rough sets as defined by~\citep{Inuiguchi}. This mathematical theory allows computing membership values that express the extent to which instances belong to each information granule~\cite{PedryczVukovich}. The intuition behind FRU is that, in fair decision-making scenarios, removing a protected feature should not cause big changes in the decision boundaries. The extent to which that happens can be used to quantify the explicit bias attached to a given protected feature.

While the FRU measure brings the added value that it does not rely on any prediction model but information granules derived from the data, it cannot capture implicit bias. For example, if a protected feature is correlated with an unprotected one, its removal might not cause significant changes to the boundary regions. This suggests that we should analyze the FRU values together with existing correlation/association patterns between protected and unprotected features. Another issue that cries for further research is the impact of fuzzy operators and distance functions on the performance of our measure.

Motivated by these two research gaps, our paper brings three main contributions. Firstly, we illustrate how the FRU is able to capture explicit bias while state-of-the-art individual-based measures struggle to capture the effect of removing a protected feature. For simulation purposes, we use the \emph{German Credit} data set~\cite{German}, which classifies loan applicants in terms of creditworthiness and is widely used in the context of AI Fairness~\cite{aif3602018}. Secondly, we conduct a sensitivity analysis to study the impact of fuzzy operators and distance functions on the FRU results. Such a study led to recommended parametric settings that can be adopted for other datasets (as reported in the supplementary materials). Finally, we discuss four scenarios that relate the changes in the boundary regions (after removing a protected feature) with the correlation/association between protected and unprotected features~\cite{prematunga2012correlational} as a way to detect implicit bias.

The remainder of the paper is organized as follows. The next section introduces the mathematical formalism behind the computation of the fuzzy-rough regions from data. Section~\ref{sec:measure} describes the similarity function we deployed and the proposed bias quantification measure. Section~\ref{sec:simulations} presents the experimental setup and analyzes the measures' outputs. Finally, Section~\ref{sec:remarks} discusses possible implications to the field.

\section{Fuzzy-rough set theory}
\label{sec:frst}

This section presents the FRS theory as described by~\citep{Inuiguchi}. This theory is used to transform tabular data into information granules characterizing each decision class. The output of this fuzzy granulation process is membership values, which will be used to define our bias quantification measure.

Let us assume that we have a universe of discourse $U$, a fuzzy set $X \in U$ and a fuzzy binary relation $R \in Q (U \times U)$ such that $\mu_{X}(x)$ and $\mu_{R}(y,x)$ are their membership functions, respectively. The membership function $\mu_X : U \rightarrow [0,1]$ determines the degree to which $x \in U$ is a member of $X$, whereas $\mu_R : U \times U \rightarrow [0,1]$ denotes the degree to which $y$ is considered to be a member of $X$ from the fact that $x$ is a member of the fuzzy set $X$. Whenever opportune, $R(x)$ is denoted with its membership function $\mu_{R(x)}(y) = \mu_{R}(y,x)$.

Firstly, let us build a partition of $U$ according to the decision classes. The $X_{k}$ set contains all objects associated with the $k$-th decision class. The membership degree of $x \in U$ to a subset $X_{k}$ was computed using the following hard membership function: $\mu_{X_{k}}(x) = 1$ for $x \in X_{k}$ and $\mu_{X_{k}}(x) = 0$ for $x \not\in X_{k}$, as we assume that all problem instances are correctly labeled.

Secondly, we need to define a fuzzy binary relation $\mu_{R}(y,x)$ to determine the fuzzy similarity between instances $x$ and $y$. This function should combine the membership degree $\mu_{X_{k}}(x)$ with the similarity degree $\phi(x,y)$ between two objects $x, y \in U$. Overall, we define $\mu_{R}(y,x) = \mu_{X_{k}}(x) \phi(x,y)$. In the next section, we will give more details about the similarity function, which is expressed in terms of a distance function.

Aiming at defining the lower approximations, we use the degree of $x$ being a member of $ X_{k}$ under the knowledge $R$. This can be measured by the truth value of the statement '$y \in R(x)$ implies $y \in X_{k}$' under fuzzy sets $R(X)$ and $X_{k}$. We use a necessity measure $inf_{y \in U} \mathcal{I}(\mu_{R}(y,x),\mu_{X_{k}}(y))$ with a fuzzy implication function $\mathcal{I}:[0,1]\times[0,1] \rightarrow [0,1]$ such that $\mathcal{I}(0,0) = \mathcal{I}(0,1) = \mathcal{I}(1,1) = 1$ and $\mathcal{I}(1,0) = 0$. It also holds that $\mathcal{I}(.,a)$ decreases and $\mathcal{I}(a,.)$ increases, $\forall a \in [0,1]$.  Equation \eqref{eqn:lower} displays the membership function for the lower approximation $R_{*}(X_{k})$ associated with the  $k$-th decision class,
\begin{equation}
\label{eqn:lower}
\mu_{R_{*}(X_{k})}(x) = min\{ \mu_{X_{k}}(x),inf_{y \in U} \mathcal{I}(\mu_{R}(y,x),\mu_{X_{k}}(y)) \}.
\end{equation}

To derive the upper approximations, we measure the truth value of the statement '$\exists y \in U$ such that $x \in R(y)$' under fuzzy sets $R(x)$ and $X_{k}$. The true value of this statement can be obtained by a possibility measure $sup_{y \in U}\mathcal{T}(\mu_{R}(x,y),$ $\mu_{X_{k}}(y))$ with a conjunction function $\mathcal{T}:[0,1]\times[0,1] \rightarrow [0,1]$ such that $\mathcal{T}(0,0) = \mathcal{T}(0,1) = \mathcal{T}(1,0) = 0$ and $\mathcal{T}(1,1) = 1$, where both $\mathcal{T}(.,a)$ and $\mathcal{T}(a,.)$ increase, $\forall a \in [0,1]$. Equation \eqref{eqn:upper} displays the membership function for the upper approximation $R^* (X_{k})$ associated with the  $k$-th decision class,
\begin{equation}
\label{eqn:upper}
\mu_{R^{*}(X_{k})}(x) = max\{ \mu_{X_{k}}(x),sup_{y \in U} \mathcal{T}(\mu_{R}(x,y),\mu_{X_{k}}(y)) \}.
\end{equation}


This model takes the minimum between $\mu_{X_{k}}(x)$ and $inf_{y \in U} $ $\mathcal{I}(\mu_{R}(y,x),\mu_{X_{k}}(y))$ when calculating $\mu_{R_{*}(X_{k})}(x)$, and the maximum between $\mu_{X_{k}}(x)$ and $sup_{y \in U}\mathcal{T}(\mu_{R}(x,y),\mu_{X_{k}}(y))$ when calculating $\mu_{R^*(X_{k})}(x)$ to preserve the inclusiveness of $R_{*}(X_{k})$ in $X_{k}$ and the inclusiveness of $X_{k}$ in $R^* (X_{k})$.

Finally, we define the fuzzy-rough regions using the upper and lower approximations. The membership functions for the fuzzy-rough positive, negative and boundary regions can be defined as $\mu_{POS(X_{k})}(x) = \mu_{R_{*}(X_{k})}(x)$, $\mu_{NEG(X_{k})}(x) = 1 - \mu_{R^{*}(X_{k})}(x)$ and $\mu_{BND(X_{k})}(x) = \mu_{R^{*}(X_{k})}(x) - \mu_{R_{*}(X_{k})}(x)$, respectively. Membership values to positive regions indicate the extent to which the instances belong to a decision class, membership values to negative regions indicate the extent to which the instances do not belong to a decision class, whereas membership values to boundary regions indicate the extent to which the instances are uncertain to the problem at hand.

\section{Fuzzy-rough uncertainty measure}
\label{sec:measure}

This section introduces our measure to quantify bias in tabular datasets used for pattern classification. This measure assumes that experts can determine the set of protected features (i.e., those likely related to bias) beforehand. The intuition of our measure is that a protected feature should not have a leading role on the decision process. For example, let us assume that we have a problem described by several features where \textit{Gender} is deemed a protected feature. If we remove that feature and there is an increase in the misclassifications, then one could conclude that \textit{Gender} is relevant to separate the decision classes. The extent to which the decision boundaries become less separate can be understood as a bias indicator.

Before presenting our measure, let us describe the similarity function \citep{FRS_survey} used to compare the instances. Such a function will be derived from a normalized heterogeneous distance function. In particular, we will employ two distance functions: the Heterogeneous Manhattan-Overlap Metric (HMOM)~\cite{Wilson1997} and the Heterogeneous Euclidean-Overlap Metric (HEOM)~\citep{Wilson1997} because of their ability to deal with instances having mixed-type features. Equation \eqref{eqn:similarity} portrays the similarity function, which produces values in the $(0,1)$ interval, 

\def\mathbi#1{\textbf{\em #1}}
\begin{equation}
\phi(x,y) = e^{-\lambda \big( d(x,y) \big)}
\label{eqn:similarity}
\end{equation}

\noindent where $\lambda > 0$ is a user-specified smoothing parameter to avoid saturation problems in which similarity values have low variability even for quite dissimilar instances.

The \textit{fuzzy-rough uncertainty} (FRU) measure \citep{koutsoviti2021bias} quantifies how much the absence of the protected feature $f_{i}$ modifies the fuzzy-rough boundary regions. If the difference is positive, we can conclude that the boundary regions became bigger after removing the protected feature, so there is more uncertainty (i.e., the feature was important for the classification). If the difference is negative, we can conclude that the boundary regions became smaller after removing the protected feature, so there is less uncertainty (i.e., the feature was causing uncertainty and its removal might be convenient).

To quantify these differences, we use the membership values of instances in $U$ to the boundary regions using (i) the full set of features, and (ii) the set of features without including the protected feature $f_{i}$ (denoted by $\neg f_{i}$). Equation \eqref{eqn:global} shows how to compute the FRU value associated with the $k$-th decision class and the $i$-th protected feature, 

\begin{equation}
\label{eqn:global}
    \Omega_{k}(f_{i}) = \frac{\sqrt{\Sigma_{x \in U}{(\Delta^{+}_{B_{k}\neg f_{i}}(x))^2}}}{\sqrt{\Sigma_{x \in U}{(\mu_{B_{k}}(x))^2}}}
\end{equation}

\noindent such that $\Delta^{+}_{B_{k}\neg f_{i}}(x) = \mu_{B_{k}}(x) - \mu_{B_{k} \neg f_{i}}(x)$ when the removal of the $i$-th feature increases the uncertainty. Otherwise, we will assume that $\Delta^{+}_{B_{k}\neg f_{i}}(x) = 0$. To lighten the notation, we denote the $k$-th boundary region $\mu_{BND(X_{k})}(x)$ with $\mu_{B_{k}}(x)$. Notice that the FRU measure is normalized by dividing by the fuzzy cardinality of the fuzzy-rough boundary region, thus leading to relative values that are not likely to be affected by class imbalance. Overall, the proposed granular measure is similar to computing the relevance of the protected feature to preserving the decision boundaries attached to the problem. Recall that in multiclass classification problems, the final FRU measure is the average FRU value of all decision classes.

\section{Numerical simulations and discussion}
\label{sec:simulations}

The case study used in our experiments is the ~\textit{German Credit} dataset, which is used for classifying loan applicants at a bank as credit worthy or the opposite. Based on the literature, protected features are \textit{Age} and \textit{Gender}~\cite{aif3602018}. \noindent Data preprocessing included (1) normalizing numeric features such that their minimum and maximum values are $0.0$ and $1.0$, respectively, (2) encoding target classes as integer identifiers starting at zero, and (3) re-coding the nominal protected feature \textit{sex\&marital status} to include only gender-related information.


Our experiments consist of four parts. The first part involves calculating the FRU values for protected features and comparing them to individual state-of-the-art measures as in~\citep{koutsoviti2021bias}. Although our paper studies the bias towards protected features, we also calculate the FRU values of unprotected features for reference. The second part is a sensitivity analysis where we examine variations in the FRU values when changing the parametric settings. In an effort to further explore the behavior of our measure, we test it on three additional datasets. The results are included in the supplementary material due to space limitations. The third part attempts to discover whether bias encoded in protected features might also be encoded in unprotected features implicitly. Finally, the last part compares our FRU measure with group-fairness measures. A separate section for such a comparison is deemed necessary as individual and group fairness measures should not be directly compared.

\subsection{Individual fairness metrics and FRU values}
\label{sec:simulations:protectedFRUliterature}

\begin{figure}[!htbp]
    \centering
    \resizebox{0.99\linewidth}{!}{
    \includegraphics{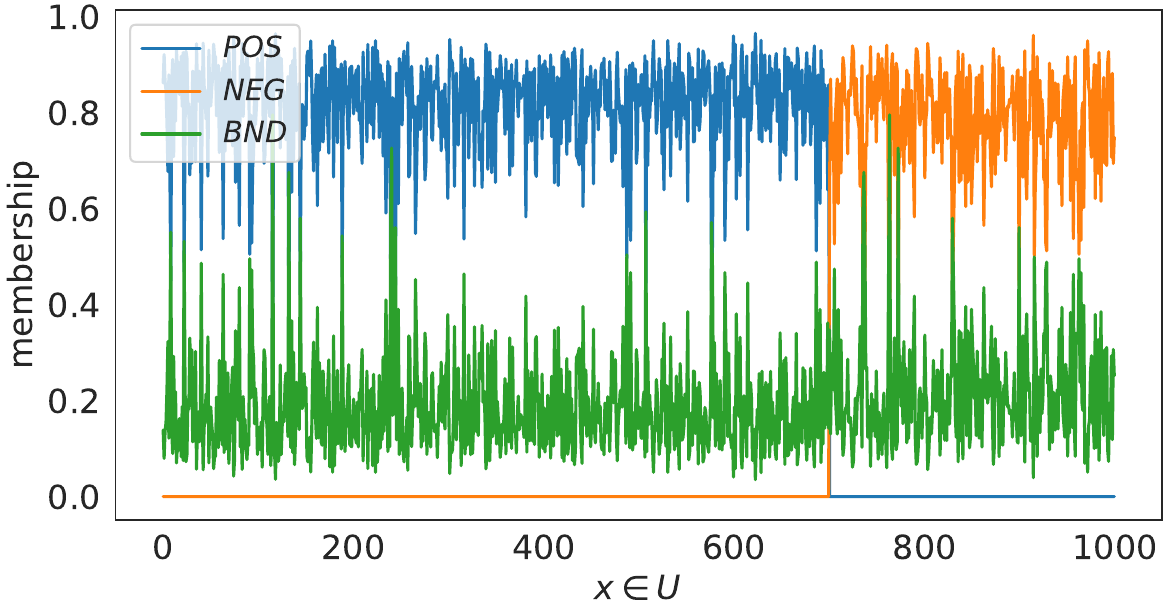}}
    \caption{Membership values to the negative, boundary and positive regions using the complete feature set. The $x$ axis represents the instances and the $y$ axis their respective membership values.}
    \label{pgf:membergerman}
\end{figure}

The first part starts with the calculation of the FRU values for protected features as in~\citep{koutsoviti2021bias}. To do that, we follow a two-step process as mentioned in Section~\ref{sec:measure}. First, the membership values to the positive, negative and boundary fuzzy-rough regions per decision class are computed using the full set of features. Figure~\ref{pgf:membergerman} shows these membership values.

\begin{figure}[!htbp]
    \centering
    \resizebox{0.99\linewidth}{!}{
    \includegraphics{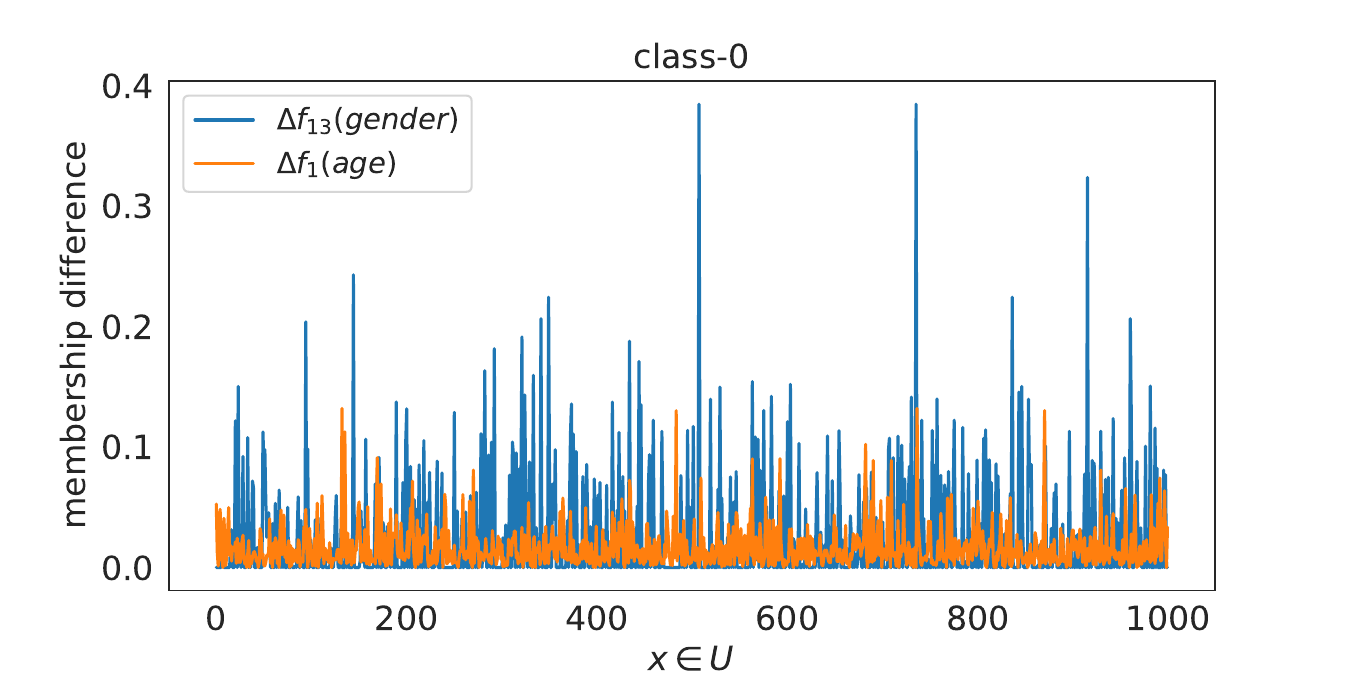}}
    \caption{Difference between membership values to the boundary regions after removing \textit{Age} ($\Delta f_{1}$) and after removing \textit{Gender} ($\Delta f_{13}$) per instance.}
    \label{pgf:boundary}
\end{figure}

The graphs show that the fuzzy-rough regions are relatively distinct from one another while involving dissimilar membership values. Second, the membership values to the regions are computed once again excluding one protected feature from the dataset. In all simulations in this sub-section, we used $\lambda=0.5$, the Łukasiewics implicator and an arbitrary t-norm. Recall that we are only interested in the positive changes that occur in the membership values to the boundary regions after suppressing a protected feature. These are used to compute the FRU values as in Equation~\eqref{eqn:global}. Figure~\ref{pgf:boundary} shows the changes in the membership values the boundary regions per instance.

Next, we compute two individual state-of-the-art measures using the aif360.sklearn package~\cite{aif3602018} and our preprocessed dataset. The first individual fairness metric is the~\textit{consistency score} (CON) coupled with a logistic regression model as the underlying predictor. The second individual metric is the~\textit{generalized entropy index} (GEI) that relies on a $k$-nearest neighbors algorithm. Let $F$ denote the set of protected and unprotected features. We are interested in exploring three settings when calculating these metrics: (i) using all features in $F$, (ii) excluding \emph{Gender} ($f_{13}$) and (iii) excluding \emph{Age} ($f_{1}$). Table~\ref{tbl:baseline:individual} shows the outputs of all measures for these settings.
\vspace{2mm}
\begin{table}[!ht]
\caption{Results of proposed and state-of-the-art measures. The ideal value of CON is one, while for the remaining ones is zero.} \label{tbl:baseline:individual} 
\centering
\resizebox{0.99\linewidth}{!}{
\begin{tabular}{|p{2cm}|p{2cm}|p{2cm}|p{2cm}|}
\hline
\multicolumn{4}{|c|}{\textbf{Individual fairness metrics}} \\ 
\hline
Feature set & CON &  GEI & FRU \\
\hline
$F$            & 0.746 & 0.093 & n/a \\
$F/\{f_{1}\}$  & 0.746 & 0.095 & 0.107  \\
$F/\{f_{13}\}$ & 0.743 & 0.093 & 0.224  \\
\hline
\end{tabular}}
\end{table}

It can be noticed that both CON and GEI  measures report roughly the same values in all three settings. The fact that the outputs of the individual fairness measures report very small changes when protected features are removed would suggest that they failed to quantify the bias issue in this problem. In contrast, our FRU measure reports larger changes in the boundary regions when the protected feature \emph{Gender} is excluded compared to~\emph{Age}. In other words, greater uncertainty in classification is reported when \textit{Gender} is suppressed. This indicates that \textit{Gender} encodes more bias than \textit{Age}. 

We continue with calculating the FRU values for all features to study the protected ones in a wider perspective. We designate this step as level-1 analysis since one feature is suppressed at a time and denote it as $\Omega(f_{i})$ where $f_{i}$ is either protected or unprotected. Observe that these values should not be interpreted as the absolute relevance of each feature in the classification process since we do not analyse the relationships for all possible feature combinations as needed in a feature selection context. Instead, we aim to investigate the extent to which protected features behave similarly to the unprotected ones. This is made possible by dividing each FRU value by the greatest FRU value among all problem features and is defined mathematically in Section~\ref{sec:simulations:case2}. Table~\ref{tab:fruncertainty} portrays these results. Moreover, we report a correlation measure between each protected feature and the unprotected ones and whether or not the correlation is significant. Further details about the correlation measure will be disclosed in the last sub-section since it will be the tool we will use to detect implicit bias in the dataset.
\vspace{2mm}
\begin{table}[!ht]
\caption{Correlation/association coefficients between protected and unprotected features, FRU values, ratio between FRU and FRU of reference.}
\resizebox{0.988\linewidth}{!}{
\begin{threeparttable}
\begin{tabular}{p{0.5cm}>{\columncolor{Gray}}p{3cm}p{1.5cm}p{1.5cm}>{\columncolor{Gray}}p{1cm}p{1cm}}
\toprule 
Idx & Features & Corr. with Gender\tnote{1} & Corr. with Age\tnote{1} &  FRU &  FRU ratio\tnote{2} \\
\midrule
\rowcolor{lightcyan}f1 & Age &   0.03* &  1.0* &     0.11 &   0.28 \\
f2 & Credit amount &  0.01* &      0.03 &     0.07 &   0.18 \\
f3 & Credit history &   0.12* &     0.03* &     0.26 &            0.67 \\
f4 & Months &    0.01 &     -0.04 &     0.11 &            0.28 \\
f5 & Foreign worker &         0.04 &       0.0 &     0.09 &            0.23 \\
f6 & Housing &        0.23* &     0.09* &     0.17 &            0.44 \\
f7 & Installment rate &         0.01 &      0.06 &      0.2 &            0.51 \\
f8 & Job &        0.09* &     0.03* &     0.23 &            0.59 \\
f9 & Existing credits &        0.01* &     0.15* &     0.09 &            0.23 \\
f10 & People liable &         0.2* &     0.01* &     0.14 &            0.36 \\
f11 &Other debtors &         0.01 &       0.0 &     0.15 &            0.38 \\
f12 & Other installment &         0.05 &       0.0 &     0.23 &   0.59 \\
\rowcolor{lightcyan} f13 & Gender &  1.0* &     0.03* &     0.22 &   0.56 \\
f14 & Employment since & 0.22* &     0.17* &     0.36 &            0.92 \\
f15 & Residence since &          0.0 &     0.27* &     0.19 &            0.49 \\
f16 & Property &  0.09* &  0.05* &     0.34 &    0.87 \\
f17 & Purpose &  0.15* &  0.03* &     0.37 &     0.95 \\
f18 & Savings account & 0.07 &      0.01 &     0.32 &  0.82 \\
\rowcolor{aliceblue} f19 & Checking account\tnote{3} & 0.03 &  0.01 &  0.39 & 1.00 \\
f20 &Telephone &        0.07* &     0.02* &     0.22 &            0.56 \\
\bottomrule
\end{tabular}
\begin{tablenotes}
\footnotesize
\item[1] Asterisks indicate significant $p$-value ($p < 0.05$) or F-statistic (larger than the critical value).
\item[2] Divide FR-Uncertainty value with FR-Uncertainty value of the reference feature
\item[3] Reference feature
\end{tablenotes}
\end{threeparttable}}
\label{tab:fruncertainty}
\end{table}

The results reveal that the feature having the largest FRU value is \textit{Checking account}, which will serve as a reference feature. We notice that \textit{Age}'s FRU value is among the five smallest FRU values and at least three times lower than \textit{Checking account}'s FRU value. \textit{Gender}'s FRU value is among the medium-ranked FRU values and about half of \textit{Checking account}'s FRU value. In this research, the change occurring in the boundary regions (as quantified by the FRU measure) after removing a protected feature is defined as \textit{explicit bias}. 

\subsection{Sensitivity analysis}
\label{sec:simulations:case2}

Next, we conduct a sensitivity analysis to measure the effect of variations in the following parameters on the measure's outputs: (1) fuzzy operators (the fuzzy implicator and the t-norm) taken from ~\citep{Napoles2018Fuzzy, Jayaram2009}, (2) the smoothing parameter in the similarity function and (3) the distance function. Tables~\ref{table:implicators} and~\ref{table:tnorms} list the different combinations to be explored.
\begin{table}[!ht]
\centering
\small
\caption{Fuzzy implicators explored in this paper.}
\label{table:implicators}
\begin{tabular}{|c|c|}
\hline
\textbf{Implicator}   & \textbf{Formulation}  \\ \hline Fodor & $\mathcal{I}_{FD}(x,y) = \begin{cases}1 & \mbox{ , } x \leq y\\\max(1-x,y) & \mbox{ , } x > y \end{cases}$ \\ \hline
Gödel & $\mathcal{I}_{GD}(x,y) = \left \{ \begin{matrix} 1 & \mbox{ , } x \leq y \\
\noalign{\smallskip} y & \mbox{ , } x > y \end{matrix} \right.$ \\ \hline
Goguen & $\mathcal{I}_{GG}(x,y) = \left \{ \begin{matrix} 1 & \mbox{ , } x \leq y \\
\noalign{\smallskip} y/x & \mbox{ , } x > y \end{matrix} \right.$ \\ \hline
Łukasiewicz & $\mathcal{I}_{LK}(x,y) = \min \{ 1-x+y,1 \}$ \\ \hline
\end{tabular}
\end{table}

\begin{table}[!ht]
\centering
\small
\caption{T-norms explored in this paper.}
\label{table:tnorms}
\begin{tabular}{|c|c|}
\hline
\textbf{T-norm}   & \textbf{Formulation}  \\ \hline
Standard intersection & $\mathcal{T}_{SI}(x,y)=\min \{x,y\}$ \\ \hline
Algebraic product & $\mathcal{T}_{AP}(x,y)=xy$ \\ \hline
Łukasiewicz & $\mathcal{T}_{LK}(x,y)=\max \{0, x+y-1 \}$ \\ \hline
Drastic product & $\mathcal{T}_{DP}(x,y) = \left \{ \begin{matrix} x & \mbox{ , } y = 1 \\
\noalign{\smallskip}
     y & \mbox{ , } x = 1  \\
\noalign{\smallskip}
     0 & \mbox{ , } otherwise \end{matrix} \right.$ \\ \hline
\end{tabular}
\end{table}

\begin{figure}[!htb]
    \centering
    \resizebox{0.99\linewidth}{!}{
    \includegraphics{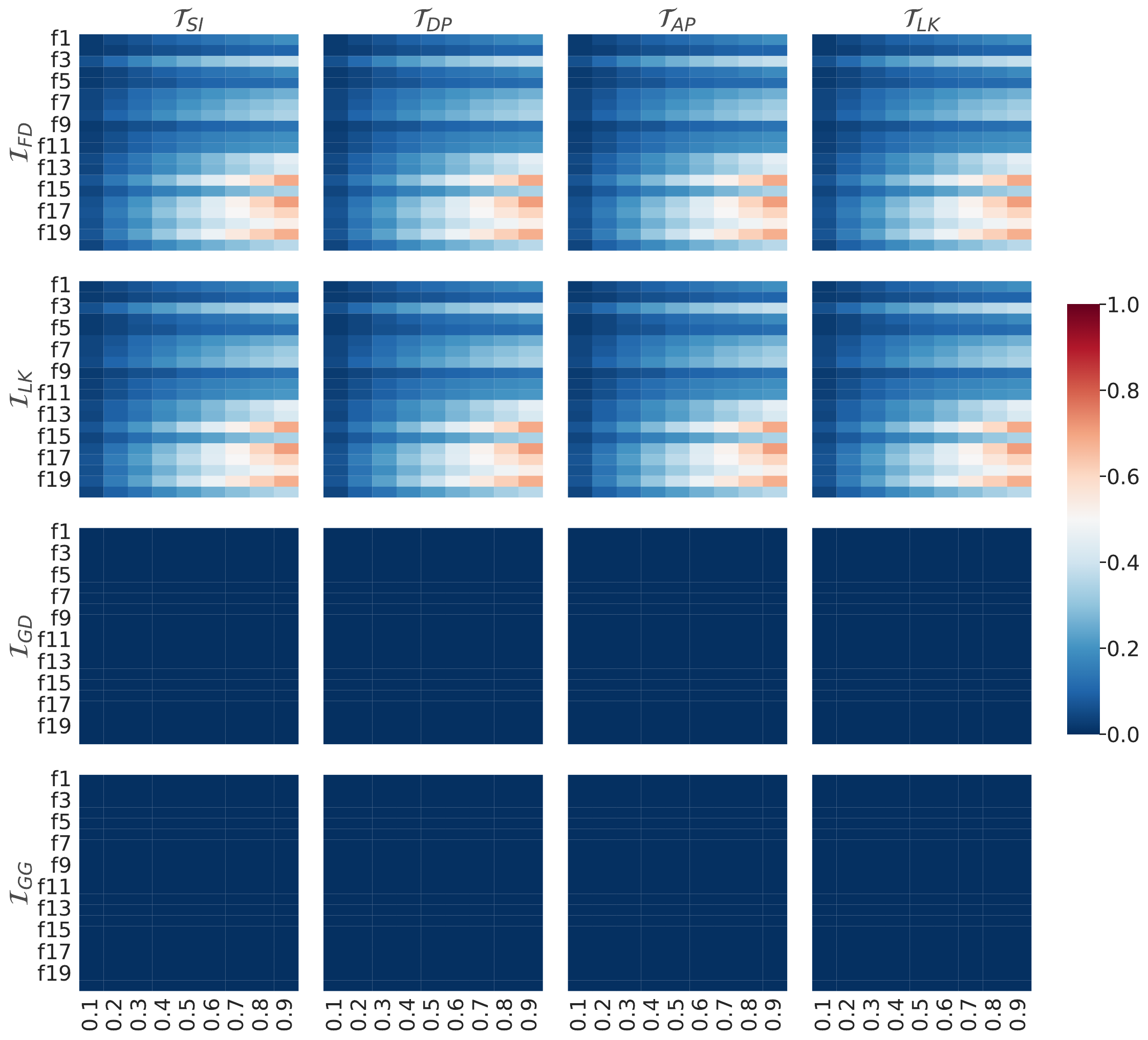}}
    \caption{Effect of the smoothing parameter ($x$ axis), fuzzy conjunction and fuzzy implicator on the FRU values. In these simulations, we use the HMOM distance function. The $y$ axis represents the problem features.}
    \label{fig:invariant HMOM}
\end{figure}

\begin{figure}[!ht]
    \centering
    \resizebox{0.99\linewidth}{!}{
    \includegraphics{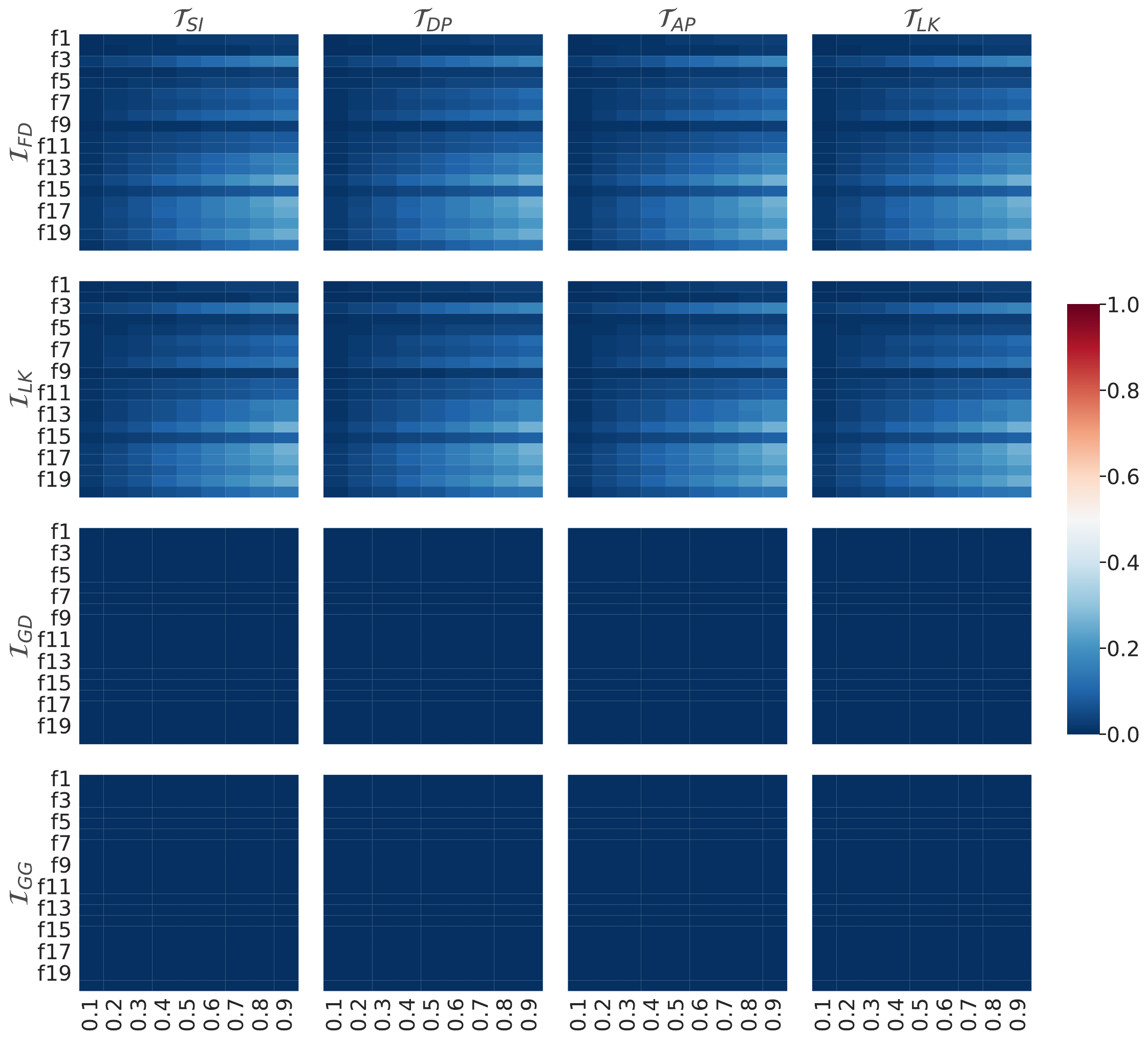}}
    \caption{Effect of the smoothing parameter ($x$ axis), fuzzy conjunction and fuzzy implicator on the FRU values. In these simulations, we use the HEOM distance function. The $y$ axis represents the problem features.}
    \label{fig:invariant HEOM}
\end{figure}

The simulation results displayed in Figures~\ref{fig:invariant HMOM} and \ref{fig:invariant HEOM} show that the choice of the fuzzy implicator has a significant impact on the measure's behavior. However, the FRU measure seems to be invariant to the choice of the fuzzy conjunction operator. Moreover, using~Łukasiewics and Fodor as the fuzzy implicator produces the same FRU values, while the rest of the implicators are unable to measure any FRU change at all~(as illustrated in the last two rows of Figures~\ref{fig:invariant HMOM} and \ref{fig:invariant HEOM}). The same patterns emerge if we use the HEOM distance function, but the changes in the FRU values ~reported by Łukasiewics and Fodor implicators are much more subtle ranging between 0.0 to 0.03. This confirms our finding that HMOM better captures changes in boundary regions. Łukasiewics is therefore chosen as the fuzzy implicator for the next round of simulations.

\begin{figure}[!htbp]
    \centering
    \begin{subfigure}[b]{0.235\textwidth}
        \includegraphics[width=\textwidth]{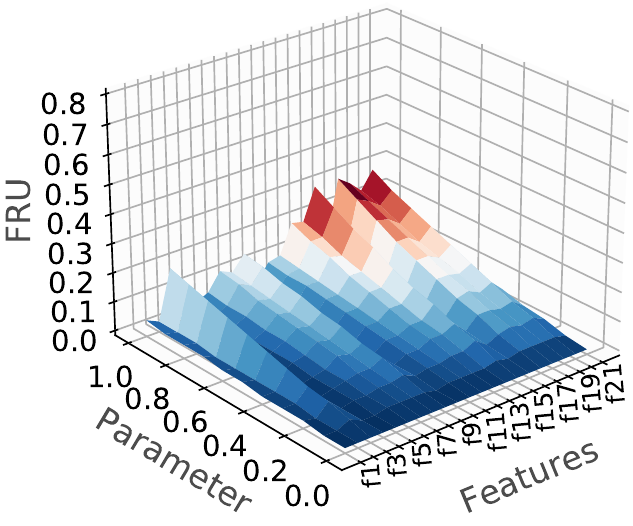}
        \caption{HEOM distance}
        \vspace{2mm}
        \label{fig:Euc_Luka_HEOM}
    \end{subfigure}
    \begin{subfigure}[b]{0.235\textwidth}
        \includegraphics[width=\textwidth]{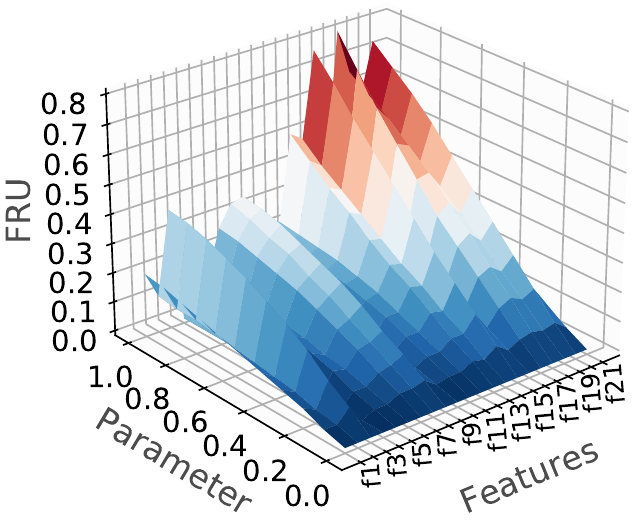}
        \caption{HMOM distance}
        \vspace{2mm}
        \label{fig:Euc_Luka_HMOM}
    \end{subfigure}    
    
\caption{Effect of the smoothing parameter and distance function on the FRU values. The measure produces larger values when using the HMOM distance function, which aligns well with the expected behavior of these distance functions. Moreover, the larger the smoothing parameter value, the larger the values produced by our measure. $\mathcal{I}_{LK}$ and $\mathcal{T}_{LK}$ are used here.}\belowcaptionskip=0ex
\label{fig:sensitivity}
\end{figure}

Figure~\ref{fig:sensitivity} offers a three-dimensional view of the FRU values at different smoothing parameter levels (from 0 to 1 with a step of 0.1) per similarity function, using Łukasiewicz both as implication and conjunction operator.

Overall, we observe that the FRU values increase when increasing the smoothing parameters. This means that increasing the smoothing parameter better separates the boundary regions. One should be careful not to confuse those variations with the absolute amount of bias measured. Therefore, we  recommend computing FRU values that are relative to the size of the boundary regions instead of using the absolute ones. These relative FRU values can be computed as $\hat{\Omega}_{k}(f_{i}) = \Omega_{k}(f_{i}) / \Omega_{k}(f_{j})$ with $f_{j}$ being a reference feature provided that $\Omega_{k}(f_{j}) > \Omega_{k}(f_{i})$. This ratio is reported in the last column of Table \ref{tab:fruncertainty}. If no reference feature is available, we can compute the relative FRU values as $\hat{\Omega}_{k}(f_{i}) = \Omega_{k}(f_{i}) / \sum_j \Omega_{k}(f_{j})$.

\subsection{Recommendations to detect implicit bias}
\label{sec:simulations:correlation}

In this subsection, we explore whether the bias encoded in the protected features explicitly, might also be implicitly encoded in unprotected features. The intuition is that implicit bias demonstrates itself when pairing two seemingly unrelated concepts~\cite{del2018conceptual, hajian2012methodology}, one of them being a protected feature. Overall, we define implicit bias as the maximal absolute correlation between the protected feature being processed and an unprotected one. Hence, we need to compute the correlation/association between each unprotected and protected feature. For the sake of simplicity, we will refer to both correlation and association as correlation unless specified otherwise. 

The correlation patterns are quantified using three different but conceptually sound statistical tools~\cite{akoglu2018user,nagelkerke1991note}. In our study, the Pearson correlation coefficient~\cite{rovine199714th} is used to measure correlation between the numeric protected feature~\textit{Age} and the rest of the numeric unprotected features. To do that, we adopt the SciPy Python package~\cite{virtanen2020scipy}. The Cram\'er's V \cite{cramer2016mathematical} is used to capture the association strength between the nominal protected feature \textit{Gender} and the unprotected nominal features. Finally, we use the R-squared coefficient of determination~\cite{nagelkerke1991note} to measure the percentage of variation in the numeric unprotected features that is explained by the protected nominal feature \textit{Gender} coupled with an F-test of joint significance~\cite{kramer2005r2}. This measure is computed using the ordinary least squares method from the statsmodels Python package~\cite{seabold2010statsmodels}. The selected measures of association are chosen to preserve consistency since they are related to the Pearson's correlation coefficient (even though features do not meet the assumptions of normality, linear dependence or homoscedasticity)~\cite{akoglu2018user}. The resulting values are reported in Table~\ref{tab:fruncertainty}. The asterisk accompanying each value represents either a $p$-value lower than 0.05 or an F-statistic larger than the critical value. In short, it refers to the confidence to which the presence or absence of correlation is observed. For example, a correlation coefficient of 0.03* should be understood as no correlation with high confidence. 

Table \ref{tab:correlation_interpretation} depicts four scenarios that can be derived from the analysis of FRU and correlation values. 

\begin{table}[!ht]
\centering
\caption{Scenarios relating correlation and FRU values.}
\resizebox{0.99\linewidth}{!}{
\begin{tabular}{|p{3.5cm}|p{3.5cm}|p{3.5cm}|}
\hline
{} & \cellcolor{Gray} Large FRU value & \cellcolor{Gray} Small FRU value \\ 
\hline
\cellcolor{Gray} Strong correlation &Explicit \& Implicit bias & Implicit bias \\ 
\hline
\cellcolor{Gray} Weak correlation & Explicit bias & Safe scenario\\
\hline
\end{tabular}}
\label{tab:correlation_interpretation}
\end{table}
\vspace{-2mm}

The scenario ``\textit{weak correlation} and \textit{large FRU value}" means that suppressing the feature causes alterations in the boundary regions. This behavior is defined as explicit bias. The scenario ``\textit{strong correlation}" and ``\textit{small FRU value}" would indicate implicit bias. In other words, the removal of the protected feature did not change the regions significantly, but the strong correlation suggests that at least an unprotected feature encodes the protected one. The scenario ``\textit{strong correlation} and \textit{large FRU value}" might imply both types of bias. Removing a protected feature that is strongly correlated with another might still cause changes in the fuzzy-rough boundary regions. It has not escaped our notice that these scenarios involve rather subjective linguistic terms such as ``\textit{strong}" or ``\textit{weak}" that should ideally be defined by domain experts.

Let us analyze a potential situation encoding implicit bias. A close inspection at the results in Table~\ref{tab:correlation_interpretation} reveals that the unprotected features \textit{Residence since} and \textit{Employment since} show the strongest correlation with \textit{Age}. While the correlation values might not be categorized as strong, it is concerning that the largest ones appear associated with unprotected features from which we can roughly infer \textit{Age}. Moreover, we notice that the feature \textit{Employment since} has the second-largest FRU value. When coupling all pieces, we can conclude that Age moderately correlates with unprotected features whose removal causes alterations in the boundary regions.

\begin{figure}[!htbp]
    \centering
    \resizebox{0.78\linewidth}{!}{
    \includegraphics{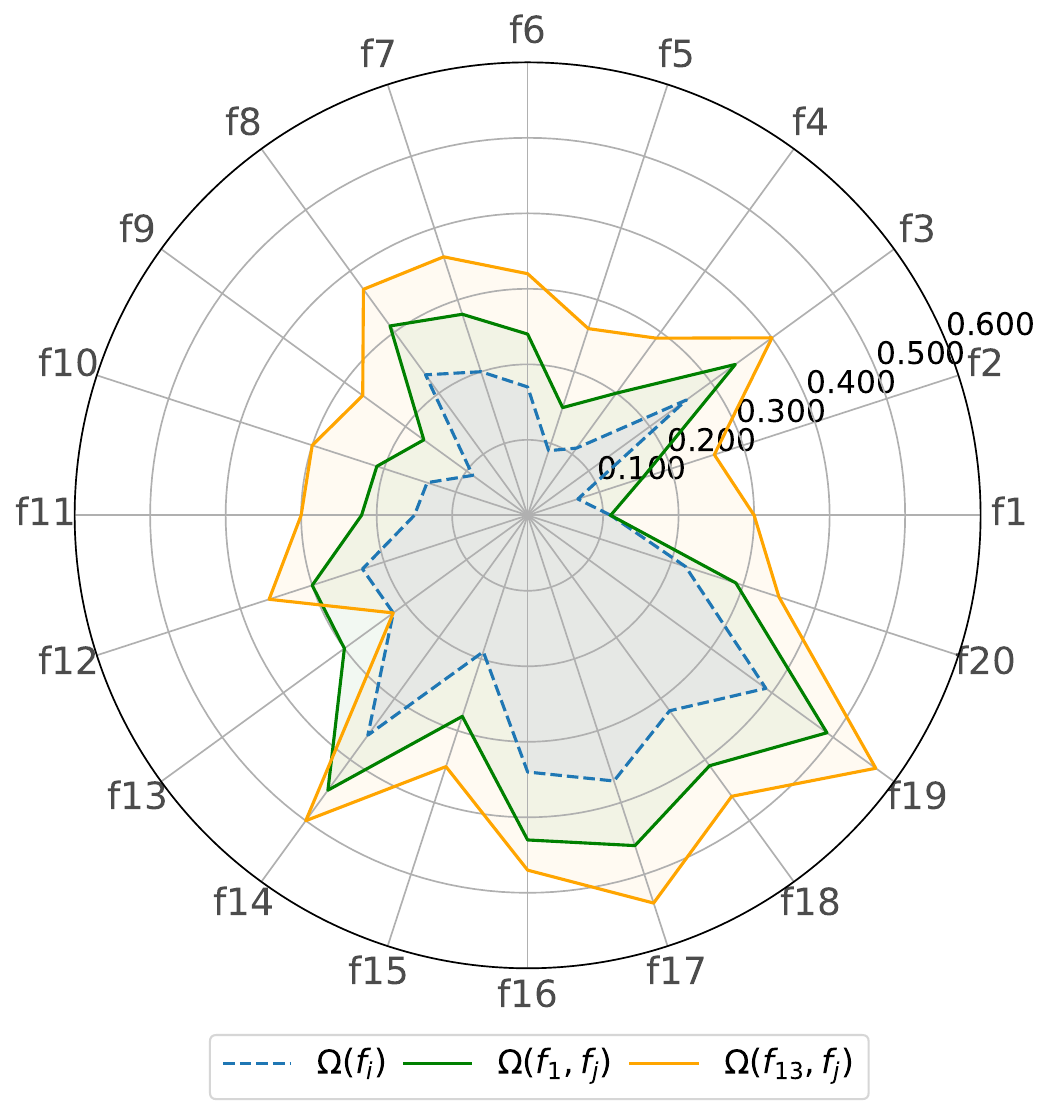}}
    \caption{FRU values when (i) suppressing each feature, (ii) suppressing each feature and \textit{Age} ($f_{1}$), and (iii) suppressing each feature and \textit{Gender} ($f_{13}$).}
    \label{fig:level2}
\end{figure}

In order to complement the analysis above, we measure the changes in the fuzzy-rough boundary regions after pairs of protected and unprotected features are excluded simultaneously. We designate this step as level-2 analysis and denote it as $\Omega(f_{i},f_{j})$ such that $f_{i}$ is a protected feature and $f_{j}$ is a unprotected one. As the level-1 analysis might not be enough to discover the role of a protected feature for the problem, we should investigate whether that same feature might become important when combined with an unprotected one. Figure~\ref{fig:level2} shows that changes in boundary regions when a single feature is excluded are relatively proportional to the changes occurring when excluded together with a protected feature.

This simulation shows that changes caused by combinations involving \textit{Gender} are larger than those involving \textit{Age}. That confirms the main finding that the results are more biased toward the former than the latter. The results also indicate that the correlation between the excluded features is not strong enough for the boundary regions to remain unchanged. However, the main conclusion from this analysis is that the binary categorization of explicit and implicit bias is too narrow: a protected feature can be important to some extent by itself when it comes to the boundary regions while also being partially encoded into unprotected features. Such a conclusion paves the road for a new research direction in which explicit and implicit biases are quantified within the fuzzy logic formalism.

\begin{figure}[!ht]
    \centering
    \resizebox{0.8\linewidth}{!}{
    \includegraphics{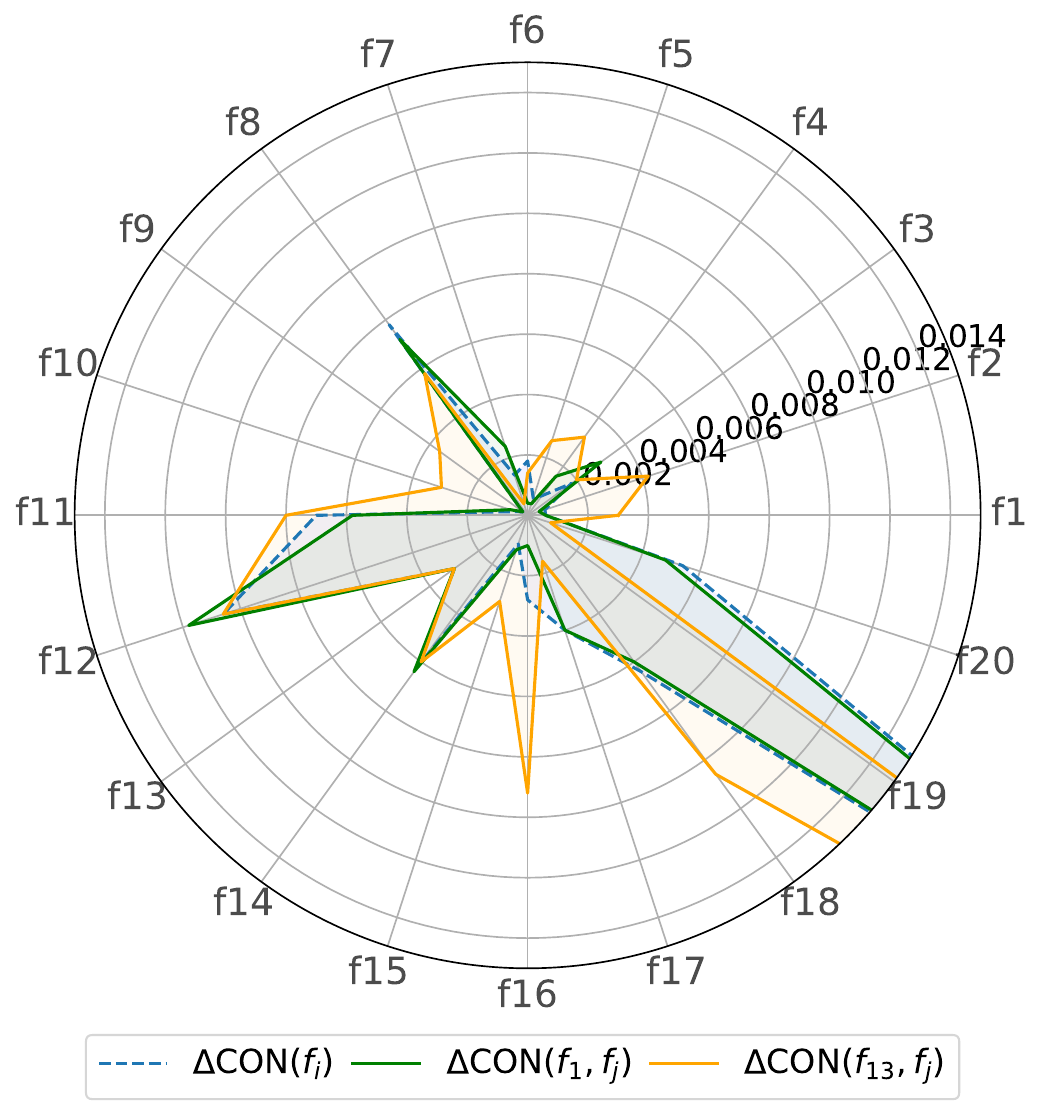}}
    \caption{$\Delta$CON values when (i) suppressing each feature, (ii) suppressing each feature and \textit{Age} ($f_{1}$), and (iii) suppressing each feature and \textit{Gender} ($f_{13}$).}
    \label{fig:level2_consi}
\end{figure}

\begin{figure}[!ht]
    \centering
    \resizebox{0.8\linewidth}{!}{
    \includegraphics{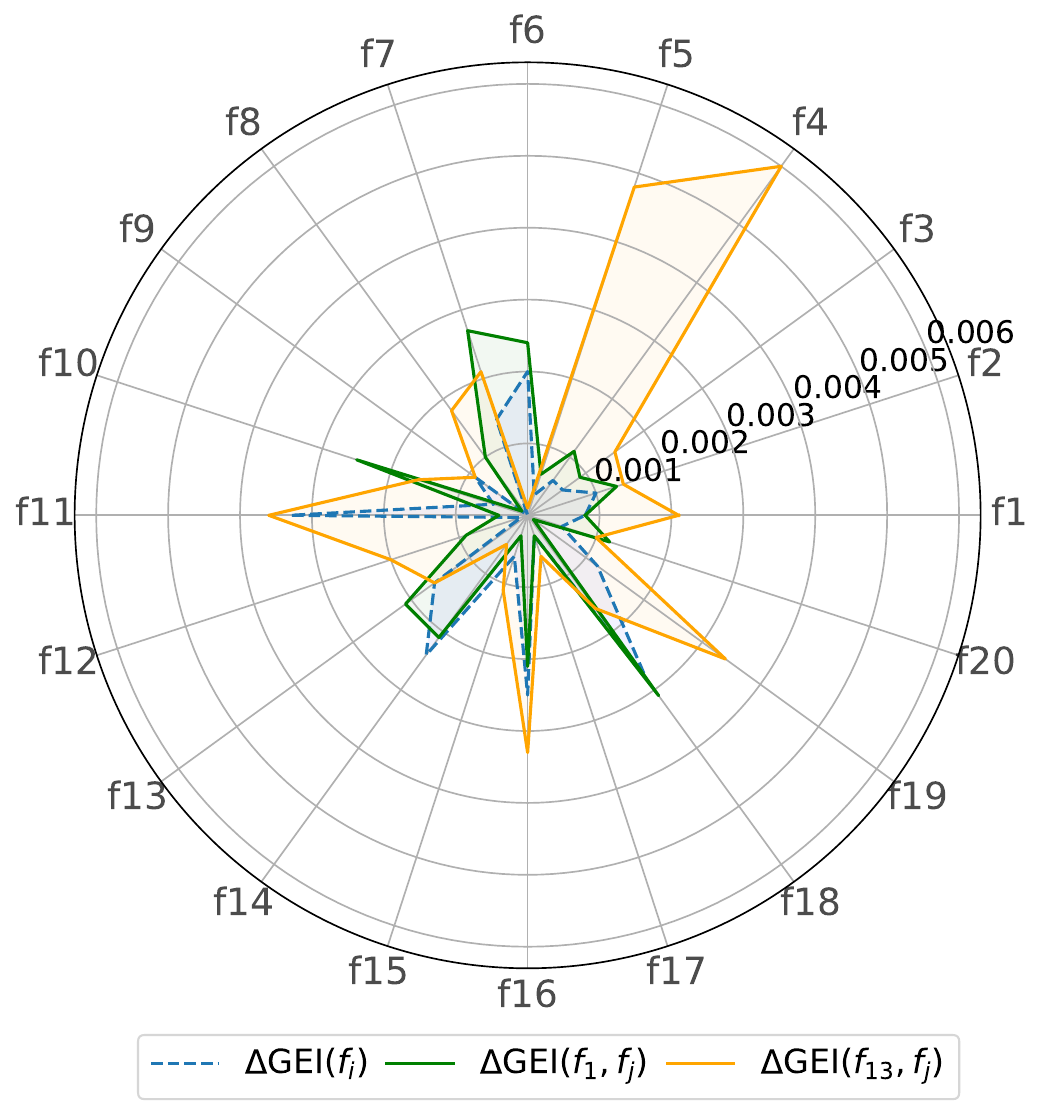}}
    \caption{$\Delta$GEI values when (i) suppressing each feature, (ii) suppressing each feature and \textit{Age} ($f_{1}$), and (iii) suppressing each feature and \textit{Gender} ($f_{13}$).}
    \label{fig:level2_GEI}
\end{figure}

Next, we compute the individual baseline measures in same the way as in the previous simulation. Let CON($F$) and GEI$(F)$ denote the values of the CON and GEI measures using the whole set of features $F$ describing the problem. Similarly, let CON$(F/\{f_i\})$ and GEI$(F/\{f_i\})$ denote the values of these measures after suppressing the protected feature $f_i$ from $F$. Finally, let CON$(F/\{f_i,f_j\})$ and GEI$(F/\{f_i,f_j\})$ be the values of these measures after removing the protected feature $f_i$ and the unprotected feature $f_j$ from $F$. The values for the level-1 analysis are computed as $\Delta$CON$(f_i)$ = $|$ CON$(F)$ - CON$(F/\{f_i\})|$ and $\Delta$GEI($f_i$) = $|$ GEI$(F)$ - GEI$(F/\{f_i\})|$. The values for the level-2 analysis are computed as $\Delta$CON($f_i,f_j$) = $|$ CON($F$) - CON$(F/\{f_i,f_j\})|$ and $\Delta$GEI($f_i,f_j$) = $|$ GEI($F$) - GEI$(F/\{f_i,f_j\})|$. We quantify the absolute difference because (i) they better illustrate the different scenarios and (ii) our FRU measure itself is the difference between the fuzzy-rough boundary regions. Figures~\ref{fig:level2_consi} and~\ref{fig:level2_GEI} compare the $\Delta$CON and $\Delta$GEI values respectively for both level-1 and level-2 analyses.

The state-of-the-art individual fairness measures report infinitesimal changes as problem features are suppressed, while the changes captured by our FRU measure vary between $0.1$ and $0.6$. Overall, these figures support our conclusion that literature measures do not capture bias in the same manner as our fuzzy-rough granulation approach.

\subsection{Comparison with group-based measures}
\label{sec:simulations:group} 

In an effort to examine bias from different perspectives, we also calculate the state-of-the-art group fairness measures using the aif360.sklearn package~\cite{aif3602018} and our preprocessed dataset. Prerequisite for computing these  measures is discretizing \textit{Age} into people younger and older than 25 years old~\cite{aif3602018}. Table~\ref{tbl:baseline:group} summarizes the outputs for the state-of-the-art measures along with the results of the FRU measure.

\begin{table}[!htb]
\vspace{3mm}
\caption{Results of proposed and state-of-the-art measures. The ideal value of \textit{Disparate Impact} is one, while for the remaining ones is zero.} \label{tbl:baseline:group} 
\centering
\resizebox{0.988\linewidth}{!}{\begin{tabular}{|p{2cm}|p{1.5cm}|p{1.5cm}|p{1.5cm}|p{1.5cm}|p{0.8cm}|}
\hline
\multicolumn{6}{|c|}{\textbf{Group fairness metrics}}\\
\hline
Protected group & Statistical parity & Disparate Impact & Equal Opportunity & Average odds & FRU\\
\hline
Gender/Female &-0.135 & 0.834 & -0.056 & -0.132 & 0.224\\
Age/Young &-0.202&	0.752&	-0.124&	-0.149 & 0.107\\
\hline
\end{tabular}}
\end{table}

Group fairness measures report slightly larger bias towards \textit{Age} than \textit{Gender}. On the contrary, our FRU measure captures the exact opposite trend which means that it is fundamentally different from existing group fairness metrics. This apparent contradiction only tells us that results might be impacted by the granularity of the bias analysis. Therefore, broader studies are often needed when analyzing bias. 

\section{Concluding remarks}
\label{sec:remarks}

This paper builds upon our recent work~\cite{koutsoviti2021bias} where we propose a measure termed \textit{fuzzy-rough uncertainty} that quantifies bias encoded in protected features. Applicable in pattern classification settings, our FRU measure quantifies the change occurring in the fuzzy-rough boundary regions after removing a protected feature. In other words, we use the change in the decision boundaries as a proxy for explicit bias. Advantages of our measure are that (i) it takes into account all features and feature categories at once, (ii) it can handle both numeric and nominal data, so no discretization is needed, (iii) it does not depend on any machine learning model to compute its outcomes but on a solid mathematical foundation, and (iv) it is less likely to be influenced by class imbalance.

The simulation results, using the ~\textit{German Credit} dataset, allow us to draw interesting conclusions. First, our measure suggests that the dataset is more biased toward \textit{Gender} than \textit{Age} when it comes to explicit bias. When contrasting our finding against the state-of-the-art measures, we observe that individual fairness measures report a barely noticeable change under the same setting. In contrast, group fairness measures show the exact opposite trend. This suggests that focusing on a particular feature-category pair instead of analyzing the protected feature as a whole might give rise to misleading results. Second, even though FRU values depend on the choice of parameters, all configurations in our sensitivity analysis consistently report greater bias against~\emph{Gender}. Third, we  recommend normalizing the FRU values of protected features using a relevant unprotected feature as reference. Moreover, we suggest using either Łukasiewicz or Fodor as implicators and the HMOM distance function since they report the largest changes. Finally, we found evidence of implicit bias in protected features (such as \textit{Age}) encoded via the unprotected features.

There are several directions to be explored in future research endeavours. Firstly, reducing the computational complexity of our algorithm is vital as it is rooted in a lazy approach that can hardly be applied if instances exceed thirty thousand. Secondly, we suggest framing the concepts of \textit{explicit bias} and \textit{implicit bias} into a multi-valued logic approach such as the fuzzy set theory. Finally, it would be convenient to analyze implicit bias taking into consideration all associations/correlations between protected and unprotected features.


\end{document}